\documentclass{article}

\usepackage[preprint,nonatbib]{neurips_2025}

\usepackage{multirow}
\newcounter{caseno}

\usepackage{xspace}
\newcommand{\ie}{\textit{i.e.}\xspace}
\usepackage{pifont}
\usepackage{amsmath}
\usepackage{booktabs}

\usepackage{makecell}
\usepackage{array}

\newcommand{\cmark}{\ding{51}}  
\newcommand{\xmark}{\ding{55}}  
\usepackage[textwidth=3.5cm]{todonotes}

\usepackage[utf8]{inputenc} 
\usepackage[T1]{fontenc}    
\usepackage{hyperref}       
\usepackage{url}            
\usepackage{booktabs}       
\usepackage{amsfonts}       
\usepackage{nicefrac}       
\usepackage{microtype}      
\usepackage{xcolor}         
\usepackage{subfigure}
\usepackage{tabularx} 
\usepackage{array}
\usepackage{ragged2e}
\usepackage{float}
\usepackage[most]{tcolorbox}
\usepackage{booktabs}
\usepackage{graphicx}
\usepackage{subcaption}  

\title{Benchmarking and Rethinking Knowledge Editing for Large Language Models}

\author{%
  Guoxiu He$^{1*}$ \quad 
  Xin Song$^{1}$\thanks{The first two authors contributed equally to this work.} \quad Futing Wang$^2$ 
  \quad Aixin Sun$^3$\thanks{Corresponding author.} \\
  $^1$East China Normal University, $^2$Westlake University, $^3$Nanyang Technological University \\
  \texttt{gxhe@fem.encu.edu.cn}, \texttt{xsong2023@stu.ecnu.edu.cn}, \\ \texttt{wangfuting@westlake.edu.cn},
  \texttt{ axsun@ntu.edu.sg} 
}

\begin{document}

\maketitle

\begin{abstract}
Knowledge editing aims to update the embedded knowledge within Large Language Models (LLMs).
However, existing approaches, whether through parameter modification or external memory integration, often suffer from inconsistent evaluation objectives and experimental setups.
To address this gap, we conduct a comprehensive benchmarking study.
In addition to fact-level datasets, we introduce more complex event-based datasets and general-purpose datasets drawn from other tasks.
Our evaluation covers both instruction-tuned and reasoning-oriented LLMs, under a realistic autoregressive inference setting rather than teacher-forced decoding.
Beyond single-edit assessments, we also evaluate multi-edit scenarios to better reflect practical demands.
We employ four evaluation dimensions, including portability, and 
compare all recent methods against a simple and straightforward baseline named Selective Contextual Reasoning (SCR). 
Empirical results reveal that parameter-based editing methods perform poorly under realistic conditions. In contrast, SCR consistently outperforms them across all settings.
This study offers new insights into the limitations of current knowledge editing methods and highlights the potential of context-based reasoning as a more robust alternative. \footnote{\url{https://github.com/ECNU-Text-Computing/Knowledge-Editing-Benchmark.git}}

\end{abstract}
\section{Introduction}
\label{sec:intro}

Large language models (LLMs) \cite{zeng2022glm, touvron2023llama, openai2023gpt} acquire extensive world knowledge \cite{jiang2020can, alkhamissi2022review, zhang2023large} and remarkable contextual reasoning abilities \cite{liu2023pre, lee2024reasoning} through large-scale pretraining \cite{brown2020language, ouyang2022training}. However, as world knowledge continuously evolves, some of the information encoded in LLMs inevitably becomes outdated or inaccurate \cite{mousavi2024your, ji2023survey}. 
Various knowledge editing methods \cite{sinitsin2020editable, zhumodifying} have been introduced to enable LLMs to incorporate updated knowledge with minimal parameter modifications or additional cost. The new knowledge pieces to be incorporated are typically collected in textual form.

\begin{figure}
    \centering
    \includegraphics[width=\textwidth]{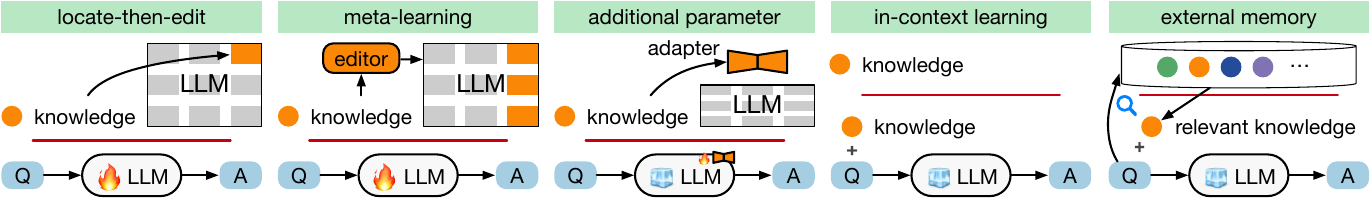}
    \caption{Training and inference workflows for five types of knowledge editing methods. The training process is shown above the \textcolor{red}{red line}; the inference stage is shown below.} 
    \label{fig:model_intro}
\end{figure}

Knowledge editing methods have evolved into five main types, distinguished by how they adjust the parameters of LLMs with reference to collected knowledge texts. Figure \ref{fig:model_intro} illustrates their respective workflows during training and inference. Specifically:
(1) \textbf{Locate-then-edit} methods \cite{meng2022locating, mengmass, li2024pmet} assume that specific knowledge is associated with certain LLM parameters. Thus, they first \textit{locate} the neurons related to the target knowledge, and then integrates new knowledge by manually \textit{editing} those parameters.
(2) \textbf{Meta-learning}-based methods \cite{de2021editing, tanmassive} assume that the patterns of parameter changes during knowledge updates can be learned. Accordingly, an editor model is trained to modify the parameters in specific layers of the LLM associated with the target knowledge.
(3) \textbf{Additional parameter}-based methods \cite{huangtransformer, yu2024melo, wang2024wise} assume that new knowledge can be stored in additional parameters external to the LLM. To achieve this, adapter layers or other auxiliary components are introduced to encode new knowledge while preserving the original parameters of the LLM.
(4) \textbf{In-context learning}-based methods \cite{zheng2023can,cohen2024evaluating} assume that the target knowledge piece relevant to a question is readily available. By embedding this knowledge directly into the prompts during inference, LLMs can utilize updated information without requiring any parameter modification.
(5) \textbf{External memory}-based methods \cite{hartvigsen2024aging, mitchell2022memory} store updated knowledge in the form of text, embeddings, hidden states, or even lightweight models. During inference, the most relevant information is retrieved to support LLM reasoning without modifying model parameters. Compared to the highly idealized in-context learning approach, external memory methods offer a more practical solution.

Despite significant progress in knowledge editing research, a unified evaluation standard has yet to be established, from knowledge format selection, to editing scenarios, then to assess dimensions. Summarized in Table \ref{tab:model_setting}, we systematically review the experimental setups of recent methods. From the perspective of \textbf{knowledge format}, most existing studies rely on fact-based knowledge (\textit{e.g.}, triplets), with limited focus on more complex event-level knowledge. Regarding \textbf{target LLMs}, current research primarily focuses on editing instruct LLMs, with little attention given to recent reasoning LLMs. 
In terms of \textbf{inference setup} for evaluation, most studies use the teacher-forcing strategy ignoring the autoregressive generation setting. 
Regarding the \textbf{editing scenarios}, many studies do not test in sequential / continuous editing scenario. 
Furthermore, in evaluating the success of knowledge editing, many studies overlook the portability \cite{zhang2024comprehensive} of the edited LLMs in real-world applications, particularly when dealing with reverse relations or multi-hop reasoning.
These evaluation gaps may obscure critical limitations, impeding a deeper understanding of the field and future advancements.

\begin{table}
  \centering
  \caption{Summary of experimental settings for 12 methods. We specify whether \textbf{Fact}-, \textbf{Event}-based, or \textbf{Gene}ral purpose \textbf{Datasets} are used. Target \textbf{LLM} specifies whether an \textbf{Instruct} LLM or an \textbf{Reason}ing LLM is employed. \textbf{Infer}ence settings show whether \textbf{Auto}regressive setting is applied instead of teacher-forcing. Under \textbf{Edits}, we clarify if \textbf{Single} or \textbf{Seq}uential editing is tested. For evaluation \textbf{Dimensions}, we specify if \textbf{Rel}iability, \textbf{Gen}eralization, \textbf{Loc}ality, and \textbf{Port}ability are considered.} 
  \scriptsize
   \resizebox{\linewidth}{!}{  
    \begin{tabular}{@{}cc|ccc|cc|c|cc|cccc@{}}

    \toprule
    \multicolumn{2}{c|}{} & \multicolumn{3}{c|}{\textbf{Datasets}} & \multicolumn{2}{c|}{\textbf{LLM}} & \textbf{Infer} & \multicolumn{2}{c|}{\textbf{Edits}} & \multicolumn{4}{c}{\textbf{Dimensions}} \\
    \textbf{Category} & \textbf{Method} & \textbf{Fact} & \textbf{Event} & \textbf{Gene.} & \textbf{Instruct} & \textbf{Reason} & \textbf{Auto} & \textbf{Single} & \textbf{Seq.} & \textbf{Rel.} & \textbf{Gen.} & \textbf{Loc.} & \textbf{Port.} \\
    \midrule
    \multirow{6}[2]{*}{Locate-then-edit} & ROME \cite{meng2022locating}  & \cmark & \textcolor{red}{\xmark} & \textcolor{red}{\xmark} & \cmark & \textcolor{red}{\xmark} & \textcolor{red}{\xmark} & \cmark & \textcolor{red}{\xmark} & \cmark & \cmark & \cmark & \textcolor{red}{\xmark} \\
          & MEMIT \cite{mengmass} & \cmark & \textcolor{red}{\xmark} & \textcolor{red}{\xmark} & \cmark & \textcolor{red}{\xmark} & \textcolor{red}{\xmark} & \cmark & \cmark & \cmark & \cmark & \cmark & \textcolor{red}{\xmark} \\
          & PMET \cite{li2024pmet}  & \cmark & \textcolor{red}{\xmark} & \textcolor{red}{\xmark} & \cmark & \textcolor{red}{\xmark} & \textcolor{red}{\xmark} & \cmark & \cmark & \cmark & \cmark & \cmark & \textcolor{red}{\xmark} \\
          & RECT \cite{gu2024model} & \cmark & \textcolor{red}{\xmark} & \textcolor{red}{\xmark} & \cmark & \textcolor{red}{\xmark} & \textcolor{red}{\xmark} & \cmark & \cmark & \cmark & \cmark & \cmark & \textcolor{red}{\xmark} \\
          & AlphaEdit \cite{fang2024alphaedit} & \cmark & \textcolor{red}{\xmark} & \cmark & \cmark & \textcolor{red}{\xmark} & \textcolor{red}{\xmark} & \cmark & \cmark & \cmark & \cmark & \cmark & \cmark \\
          & FT-L \cite{meng2022locating}  & \cmark & \textcolor{red}{\xmark} & \textcolor{red}{\xmark} & \cmark & \textcolor{red}{\xmark} & \textcolor{red}{\xmark} & \cmark & \textcolor{red}{\xmark} & \cmark & \cmark & \cmark & \textcolor{red}{\xmark} \\
    \midrule
    Meta-learning & MEND \cite{de2021editing}  & \cmark & \textcolor{red}{\xmark} & \textcolor{red}{\xmark} & \cmark & \textcolor{red}{\xmark} & \textcolor{red}{\xmark} & \cmark & \textcolor{red}{\xmark} & \cmark & \cmark & \cmark & \textcolor{red}{\xmark} \\
    \midrule
    Additional & AdaLoRA \cite{zhang2023adalora} & \cmark & \textcolor{red}{\xmark} & \textcolor{red}{\xmark} & \cmark & \textcolor{red}{\xmark} & \textcolor{red}{\xmark} & \cmark & \textcolor{red}{\xmark} & \cmark & \cmark & \cmark & \textcolor{red}{\xmark} \\
    Parameter & WISE \cite{wang2024wise} & \cmark & \textcolor{red}{\xmark} & \textcolor{red}{\xmark} & \cmark & \textcolor{red}{\xmark} & \textcolor{red}{\xmark} & \cmark & \cmark & \cmark & \cmark & \cmark & \textcolor{red}{\xmark} \\
    \midrule
    In-context & IKE \cite{zheng2023can}  & \cmark & \textcolor{red}{\xmark} & \textcolor{red}{\xmark} & \cmark & \textcolor{red}{\xmark} & \textcolor{red}{\xmark} & \cmark & \cmark & \cmark & \cmark & \cmark & \textcolor{red}{\xmark} \\
    Learning & ICE \cite{cohen2024evaluating}  & \cmark & \textcolor{red}{\xmark} & \textcolor{red}{\xmark} & \cmark & \textcolor{red}{\xmark} & \cmark & \cmark & \textcolor{red}{\xmark} & \cmark & \cmark & \cmark & \cmark \\
    \midrule
    External Memory & GRACE \cite{hartvigsen2024aging} & \cmark & \textcolor{red}{\xmark} & \textcolor{red}{\xmark} & \cmark & \textcolor{red}{\xmark} & \textcolor{red}{\xmark} & \cmark & \cmark & \cmark & \cmark & \cmark & \textcolor{red}{\xmark} \\
    \bottomrule
    \end{tabular}%
    }
  \label{tab:model_setting}%
\end{table}%

To this end, we conduct comprehensive benchmarking experiments that simulate practical knowledge editing scenarios. We recognize that real-world applications often require multiple sequential edits as knowledge evolves over time. We adopt a unified autoregressive inference setting, as edited LLMs should not rely on ground truth access as in teacher-forcing settings. Using instruct LLMs as our foundation, we perform knowledge editing with factual data and evaluate performance across four dimensions. Beyond standard single-edit evaluations, we assess the methods' capability to handle sequential editing scenarios. We systematically compare these effects on both instruct and reasoning-specialized LLMs, while also evaluating the preservation of mathematical reasoning abilities post-editing. To complement our quantitative analysis, we conduct detailed case studies and error analyses. Notably, we introduce Selective Contextual Reasoning (SCR), a straightforward baseline that retrieves relevant knowledge in response to queries and incorporates it as contextual prompts during generation. Comparisons with mainstream knowledge editing methods provide intuitive insights into their strengths and limitations.

Our main findings are as follows: 
(1) Most knowledge editing methods that rely on parameter modification exhibit poor transferability and largely lose their multi-hop reasoning capabilities. 
(2) As the number of edits increases, the performance of these methods degrades sharply, rendering them impractical for real-world use.
(3) After editing, LLMs tend to suffer significant declines in performance, including the loss of basic reasoning and processing abilities.
(4) In contrast, knowledge editing methods that do not require parameter modification such as SCR demonstrate greater stability, highlighting the limitations of assumptions linking parameters to specific knowledge.
\section{Related Work}
\label{sec:related}

This paper is dedicated to revisiting knowledge updating in LLMs. We review existing knowledge editing methods, followed by a critical analysis of concerns associated with these methods.

\subsection{Knowledge Editing Methods}

LLMs are often regarded as knowledge bases, as they encapsulate vast amounts of world knowledge within their extensive parameters, which are derived from large-scale datasets during the pre-training phase~\cite{petroni2019language, geva2021transformer, geva2023dissecting, dai2022knowledge,wangknowledge}. 
To cope with knowledge updates, the knowledge editing approach~\cite{meng2022locating, mengmass} encodes target knowledge into specific parameters, which are then replaced or supplemented in the LLM to update its factual knowledge.

\textbf{Locate-then-edit} methods~\cite{mengmass, zhang2024knowledge, li2024pmet, li2024consecutive,hu2024wilke} are ground in the interpretability theory of Transformer architecture~\cite{geva2021transformer,lv2024interpreting}. It posits that knowledge is distributed across feed-forward networks (FFNs), while attention modules play a role in information copying and transmission. For instance, ROME~\cite{meng2022locating} utilizes causal tracing to first identify the crucial neurons associated with specific knowledge before performing targeted edits. 
Furthermore, RECT~\cite{gu2024model} and AlphaEdit~\cite{fang2024alphaedit} introduce additional constraints based on ROME to prevent excessive parameter shifts during the editing process.
Note that, the assumption of the localized storage of factual knowledge remains controversial~\cite{wei2024does}. 
An alternative hypothesis suggests that the relationship between neurons and knowledge is characterized by a many-to-many dynamic rather than a simplistic one-to-one association~\cite{allen2024physics}. Any modification to the parameters will inevitably affect other knowledge stored in the LLM, including both the original knowledge, and previously edited knowledge.

\textbf{Meta-learning} methods, such as MEND~\cite{de2021editing} and MALMEN~\cite{tanmassive}, employ hyper-networks that are designed to forecast tailored weight updates for each knowledge data instance associated with an LLM. However, the hyper-network for a particular LLM limits their scalability in sequential editing scenarios. Furthermore, the additional training process incurs significant time and computational costs. Additionally, the necessity of modifying parameters for a limited amount of knowledge encapsulated in textual form is a matter of ongoing debate.

\textbf{Additional parameter-based} methods aim to efficiently integrate target knowledge by isolating the parameters that require adjustment, such as WISE~\cite{wang2024wise}, T-Patcher~\cite{huangtransformer}, MELO~\cite{yu2024melo} and many others~\cite{dong2022calibrating, zhang2024dafnet, wang2024memoe, wang2024lemoe, wang2024roselora}. 
These methods typically introduce additional parameters or utilize mixture of experts (MoE) architectures~\cite{chen2022towards}, either at the head of the LLM or within its structure.
However, as the number of additional parameters increases, the likelihood of overfitting escalates. This phenomenon can result in the edited LLM neglecting prior edits, and compromising its original knowledge. Besides, the continual expansion of neurons may further exacerbate the edited LLM's inference burden.

\textbf{In-context learning}-based methods \cite{zheng2023can,cohen2024evaluating} assume that the target knowledge piece relevant to a question is readily available. By embedding this knowledge directly into the prompts during inference, LLMs can utilize updated information without requiring any parameter modification. IKE~\cite{zheng2023can} enables the LLM to learn copy, update, and retain information through examples.
Note that, these methods typically require frequent updates to the classifier as the memory expands, necessitating continuous training of the additional discriminator models.
In the deeper layers of large-scale neural models, most feature values diminish significantly, compressing the representational space into a limited set of feature directions. This phenomenon, known as dimensional collapse~\cite{dohmatob2024strong}, undermines the reliability of hidden states for encoding and retaining edited knowledge. And IKE~\cite{zheng2023can} places greater emphasis on selecting contextual demonstrations. Incorporating relevant updated knowledge directly into the prompt, without retrieval, does not align with real-world scenarios.

\textbf{External Memory-based} methods~\cite{zhong2023mquake,hartvigsen2024aging,mitchell2022memory,jiang2024learning, chen2024lifelong, daslarimar, zheng2023can, markowitz2025k} maintain a memory store for updated knowledge, which can be represented as plain text, hidden states, token embeddings, or knowledge graphs. SERAC~\cite{mitchell2022memory}, a classical method, simulates the editing scope by training a discriminator, whose results distinguish between the original LLM and the counterfactual model. GRACE~\cite{hartvigsen2024aging} maintains a dynamically updated codebook that alters the hidden states during the forward propagation.

In short, most existing methods update LLM knowledge by modifying their parameters or structures. However, in real-world scenarios, the amount of new knowledge available is limited, compared to the vast pre-training data used by LLMs. Encoding new knowledge into the LLM's parameters can lead to the loss of original knowledge, risking both an incomplete understanding of updates and potential conflicts with prior knowledge.

\subsection{Critiques of Knowledge Editing}

Knowledge editing allows for targeted modifications to an LLM's knowledge, but many studies \cite{hsueh2024editing, pinter2023emptying, wang2024missing, gu2024model, yang2024butterfly, yang2024fall} suggest it may also introduce negative consequences. 
\cite{pinter2023emptying} criticize that direct knowledge editing pursues factual accuracy, which is misaligned with the pre-training objectives of LLMs. They caution that knowledge editing reinforces the flawed notion that \textit{model authenticity is reliable} and cannot serve as a remedy for the inherent shortcomings of LLMs. 
\cite{wang2024missing} quantitatively evaluates the negative impact of the ripple effect in the hidden space, which significant hinders the effectiveness of editing tasks and overall performance of edited LLMs. 
Similarly, \cite{gu2024model} demonstrates that improvements in factuality come at the cost of a significant decline in reasoning, natural language inference, and question-answering abilities. LLMs are not robust to parameter perturbations, as editing 1\% of parameters can cause a sharp decline in other tasks. Achieving both factual accuracy and general capability remains a challenging dual objective. 
\cite{yang2024butterfly} also experimentally confirms that after a few edits, the coherence of the edited LLM's text generation decreased significantly, leading its performance on downstream tasks to approximate random guessing. 
Additionally, \cite{yang2024fall} demonstrates that ROME could cause the LLM to crash with just a single edit. 
And \cite{halevy2024flex} finds that editing model parameters can exacerbate existing biases against certain demographic groups and amplify misinformation, intensifying issues like racial biases and gender discrimination to varying degree across all methods and models. 

These studies indicate that direct knowledge editing is neither an effective nor a reliable solution for addressing knowledge obsolescence in LLMs. 
However, these studies lack unified evaluation datasets, objectives, and standards, which limits comprehensive assessment and deeper understanding of the knowledge editing, as shown in Table \ref{tab:model_setting}. In this paper, we categorize knowledge editing methods into five types: three involving parameter modifications and two that do not. We then consider more complex and general datasets, more realistic editing frequency and inference settings, account for all key performance dimensions, and introduce the most intuitive baseline method. We aim to conduct a comprehensive benchmark study to uncover potential oversights in critical application-relevant characteristics.

\section{Preliminaries}  
\label{sec:preliminaries}

The goal of triplet-based knowledge editing algorithm $\mathcal{E}$ is updating $e=(h, r, t)$ to $e'=(h, r, t_*)$, \textit{e.g.}, (Barack Obama, born in, Hawaii) to (Barack Obama, born in, Kenya), within original LLM $f^0$, the edited LLM $f^1$ can be got as follows:
\begin{equation}
   f^1 = \mathcal{E}(f^0, e'), \quad \text{such that} \ e' \in f^1
\end{equation}
 
Factual knowledge, such as triples, is represented as a single prompt–answer pair $e=[(x, y)]$, whereas more complex event is represented using multiple prompt–answer pairs $e = [(x_1, y_1), \cdots, (x_m, y_m)]$, where $x$ denotes the prompt to elicit the knowledge, $y$ represents the answer, and $m$ denotes the number of pairs. After editing the LLM $f^0$, it is expected that, given a knowledge-related prompt $x$, the LLM $f^1$ will produce the correct answer $y$.

\paragraph{Editing Scenarios:} \textbf{Single Editing} refers to the process where the LLM updates only one piece of knowledge ($E = [(e_1)]$) at a time.
\textbf{Sequential Editing} also known as lifelong editing or continual editing \cite{hartvigsen2024aging,yu2024melo,wang2024wise}, refers to repeatedly applying the editing method to update a sequence of knowledge $E_t = [e_1, e_2, \ldots, e_t]$.
Consequently, the LLM evolves from $f^0$ to $f^t$ over multiple updates. Note that, in the sequential editing setting, each method operates differently: \textbf{locate-then-edit} methods modify a subset of the LLM's parameters with each update; \textbf{meta-learning}-based methods continuously update the parameters of the dedicated editor;
 \textbf{additional parameter}-based methods repeatedly update external parameter components; \textbf{in-context learning}-based methods always provide the most relevant knowledge during inference, regardless of the number of updates, an idealized scenario; \textbf{external memory}-based methods only update an external knowledge base, such as a text corpus or vector store, that stores newly introduced knowledge.

\paragraph{Evaluation Dimensions:} After incorporating $t$ pieces of new knowledge, \ie undergoing $t$ rounds of updates, the edited LLM $f^t$ is evaluated in four dimensions~\cite{zhang2024comprehensive}:

\begin{itemize}
    \item \textbf{Reliability:} The edited LLM should generate the updated target output for the prompts used for knowledge editing, \ie those available in $E_t$,  and ensure the persistence of the editing effects. This can be formally expressed as:
    \begin{equation}
        \mathbb{E}_{(x_i, y_i) \sim E_t} \mathbb{I} \{ \arg\max_y f^t(y \mid x_i) = y_i \}.
    \end{equation}

    \item \textbf{Generalization:} The edited LLM should extend beyond the exact edits and correctly respond to paraphrased prompts, denoted as $N(E_t)$. Mathematically, this is formulated as:  
    \begin{equation}
        \mathbb{E}_{(x_i, y_i) \sim N(E_t)} \mathbb{I} \{ \arg\max_y f^t(y \mid x_i) = y_i \}.
    \end{equation}

    \item \textbf{Locality:} It is imperative that the edited LLM should retain its original behavior when processing queries that are unrelated to the edits, denoted by $O(E_t)$. This criterion measures how well the edited LLM maintains the overall stability while keeping edits confined to the relevant scope. This requirement can be represented as:
    \begin{equation}
        \mathbb{E}_{(x_i, y_i) \sim O(E_t)} \mathbb{I} \{ f^t(y \mid x_i) = f^0(y \mid x_i) \}.
    \end{equation}

    \item \textbf{Portability:} Furthermore, the edited LLM should effectively propagate the impact of the edited knowledge, correctly reasoning about its downstream implications, denoted by $D(E_t)$. $D(E_t)$ encompasses three aspects: substituting the subject of the question with aliases, reasoning based on factual changes, and knowledge derived from reverse relationships. As shown in Table \ref{tab:model_setting}, this dimension has been ignored by a large number of studies. The requirement is defined as:
    \begin{equation}
        \mathbb{E}_{(x_i, y_i) \sim D(E_t)} \mathbb{I} \{ \arg\max_y f^t(y \mid x_i) = y_i \}.
    \end{equation}
\end{itemize}

\paragraph{Evaluation for General Tasks:} In addition to the knowledge editing evaluation dimensions, it is also important to assess whether the edited LLM retains its ability to handle general tasks. In this study, we use a mathematical reasoning dataset to evaluate the inference performance of the edited LLM on such tasks. 

If the edited LLM adheres to these requirements, then the updating process is considered robust, precise, and minimally disruptive to unrelated LLM behavior. 

\section{Experiments and Results}
\label{sec:experiments}

\subsection{Experimental Settings}

\paragraph{LLMs} We primarily conduct experiments on decoder-only LLMs, focusing specifically on Llama-2-7B-Chat \cite{touvron2023llama}, Llama-3.1-8B-Instruct \cite{llama3modelcard}, and Mistral-7B-Instruct \cite{jiang2023mistral}. In addition to these general-purpose LLMs, we perform knowledge editing on DeepSeek-R1-Distill-LLaMA-8B \cite{guo2025deepseek} to examine the effectiveness and potential impact of current editing methods on reasoning-oriented LLMs.

\paragraph{Knowledge Editing Methods} This study examines twelve recent knowledge editing methods. Among them, methods involving parameter modification include locate-then-edit methods such as ROME \cite{meng2022locating}, MEMIT \cite{mengmass}, PMET \cite{li2024pmet}, RECT \cite{gu2024model}, AlphaEdit \cite{fang2024alphaedit}, and FT-L \cite{meng2022locating}; meta-learning-based methods such as MEND \cite{de2021editing}; and additional-parameter-based methods such as AdaLoRA \cite{zhang2023adalora} and WISE \cite{wang2024wise}. In contrast, methods that do not involve parameter modification include in-context learning-based methods such as IKE \cite{zheng2023can} and ICE \cite{cohen2024evaluating}, as well as external memory-based methods such as GRACE \cite{hartvigsen2024aging}.

\textbf{An Intuitive Baseline} We introduce a simple and intuitive baseline called Selective Contextual Reasoning (SCR) \cite{he2025knowledge}, which falls under the category of external memory-based knowledge editing method.\footnote{This work extends \cite{he2025knowledge} by providing a comprehensive benchmark, with broader coverage of datasets, LLM models, and measures.} This method stores all knowledge in textual form. When faced with a new question, it first retrieves the most relevant knowledge from the knowledge base. If relevant knowledge is found, it is included in the prompt along with the question for the LLM to answer. If not, the LLM answers based solely on the question.  For detailed implementation, see the Appendix \ref{scr}.

\paragraph{Knowledge Editing Datasets} We use two widely adopted context-free question answering (QA) datasets in knowledge editing research: \textbf{WikiData\(_\text{counterfact}\)}\cite{cohen2024evaluating} and Zero-Shot Relation Extraction (\textbf{ZsRE})\cite{levy2017zero}, both of which serve as counterfactual benchmarks for modifying knowledge in LLMs. In addition, we incorporate an event-centric dataset, \textbf{ELKEN} \cite{peng2024event}, in which each instance contains multiple related statements, to evaluate the ability of knowledge editing methods to handle more abundant and complex knowledge.
\begin{itemize}
\item \textbf{ZsRE}: Originally a question-answering dataset, ZsRE is extended by \cite{yao2023editing} to assess various dimensions of knowledge editing methods. 

\item \textbf{WikiData\(_\text{counterfact}\)}: This dataset collects triplets about popular entities, ensuring that the subject corresponds to one of the most viewed Wikipedia pages. 

\item \textbf{ELKEN}: This dataset is designed for event-level knowledge editing, focusing on directly editing new events into LLMs. It includes a diverse set of events and corresponding questions about factual knowledge and future trends.  Due to the fact that tendency predictions are influenced by factors beyond the current edited event, we have filtered out questions related to future trends, focusing exclusively on factual knowledge. 
\end{itemize}

The first two datasets, ZsRE and WikiData\(_\text{counterfact}\), are from KnowEdit\footnote{\url{https://huggingface.co/datasets/zjunlp/KnowEdit}} and released under the \textbf{MIT License}. The ELKEN dataset\footnote{\url{https://github.com/THU-KEG/Event-Level-Knowledge-Editing.git}} follows the \textbf{CC BY-NC-SA 4.0} license.

\paragraph{General Datasets} To evaluate the ability of edited LLMs to handle general tasks, we specifically select mathematical reasoning benchmarks to assess the extent to which their reasoning capabilities are preserved after editing. The selected benchmarks include AIME 2024, AIME 2025, AMC \cite{li2024numinamath}, OlympiadBench \cite{he2024olympiadbench}, and MATH-500 \cite{hendrycks2021measuring}.

\paragraph{Inference Setting of Knowledge Editing} Following methods such as MEND \cite{mitchell2021fast} and ROME \cite{melocang2022locating}, several follow-up studies \cite{zhang2024comprehensive,wang2024wise} have adopted teacher forcing \cite{williams1989learning} during inference and evaluated token-level changes. While this evaluation setting is widely used, it relies on ground-truth answer sequences at inference time, which introduces potential data leakage. Such setups may lead to overly optimistic results and fail to accurately reflect the method's effectiveness in real-world generative scenarios, as also highlighted by \cite{yang2025mirage}. To ensure a more fair and realistic evaluation, we adopt a unified autoregressive generation paradigm \cite{mccoy2023embers} for prediction. We directly evaluate whether the generated answer is correct with respect to the target answer.

\paragraph{Sequential Editing Scenario} Since single editing is less representative of real-world applications, we conduct experiments under a sequential editing scenario. For each dataset, we edit each knowledge item in sequence, treating the final LLM as the fully edited version. 

\paragraph{Event Knowledge Editing Settings} For parameter-modification-based knowledge editing methods, we use GPT-4o \cite{hurst2024gpt} to convert each event into a set of fact triples, which are then used for sequential editing. The prompt used is in Appendix \ref{prompt}. For context-based and external memory-based methods, the original event text is directly provided as context input or integrated into the editable memory.

\paragraph{Experimental Environment}
The experiments are executed on 8 NVIDIA A800
GPUs under a Linux system. All methods are implemented using EasyEdit \cite{wang2024easyedit}.\footnote{\url{https://github.com/zjunlp/EasyEdit.git}} Evaluation on reasoning benchmarks is conducted using LUFFY \cite{yan2025learning}.\footnote{\url{https://github.com/ElliottYan/LUFFY.git}}

\paragraph{Metrics Calculation} For the instruct LLM, we limit the max output token length to 50, while for the reasoning LLM, it is set to 1024. For the four evaluation dimensions, we use Qwen2.5-72B-Instruct to assess whether the answers generated by the edited LLMs are semantically consistent with the ground-truth answers, and compute their accuracy (\%) accordingly. 
To evaluate the impact of different editing loads on the performance of knowledge editing methods, we test scenarios involving 1, 10, and 100 edits. Specifically, we select the first 100 knowledge items from the dataset. For the 1-edit setting, each item is edited individually, and the average performance across all 100 edits is reported. For the 10-edit setting, the 100 items are evenly divided into 10 groups, with sequential editing applied within each group; the average performance after each 10-edit sequence is then used. For the 100-edit setting, we sequentially edit all 100 items. 

\subsection{Results and Analysis}
 
\textbf{RQ1: How do knowledge editing methods perform in practice-oriented settings?}

\begin{figure}
    \centering

    \begin{subfigure}
        \centering
        \includegraphics[width=0.9\textwidth]{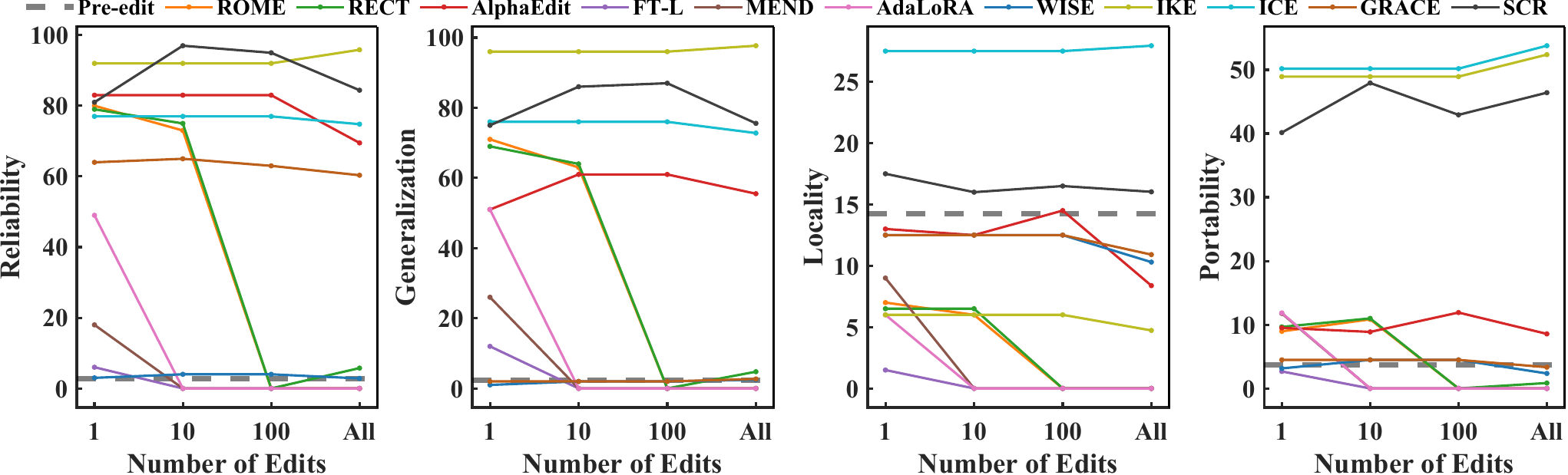}
        \caption*{(a) LLaMA-3.1-8B-Instruct}
        \label{fig:scale_llama3}
    \end{subfigure}
    
    \vspace{1em}
    
    \begin{subfigure}
        \centering
        \includegraphics[width=0.95\textwidth]{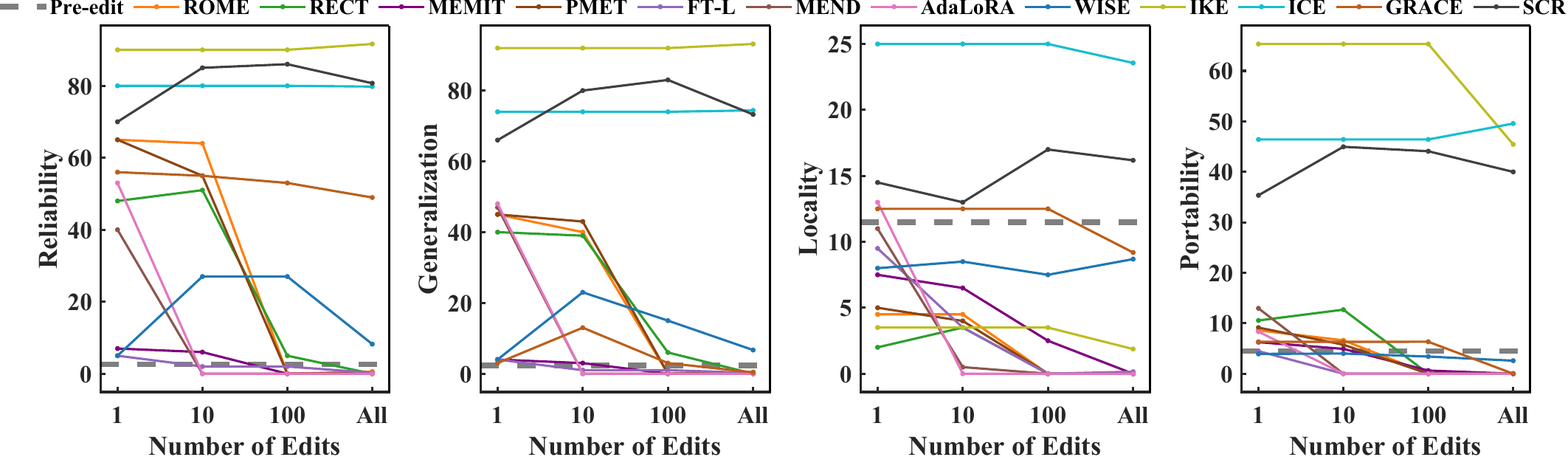}
        \caption*{(b) LLaMA-2-7B-Chat}
        \label{fig:scale_llama2}
    \end{subfigure}
    
    \vspace{1em}
    
    \begin{subfigure}
    \centering
        \includegraphics[width=0.95\textwidth]{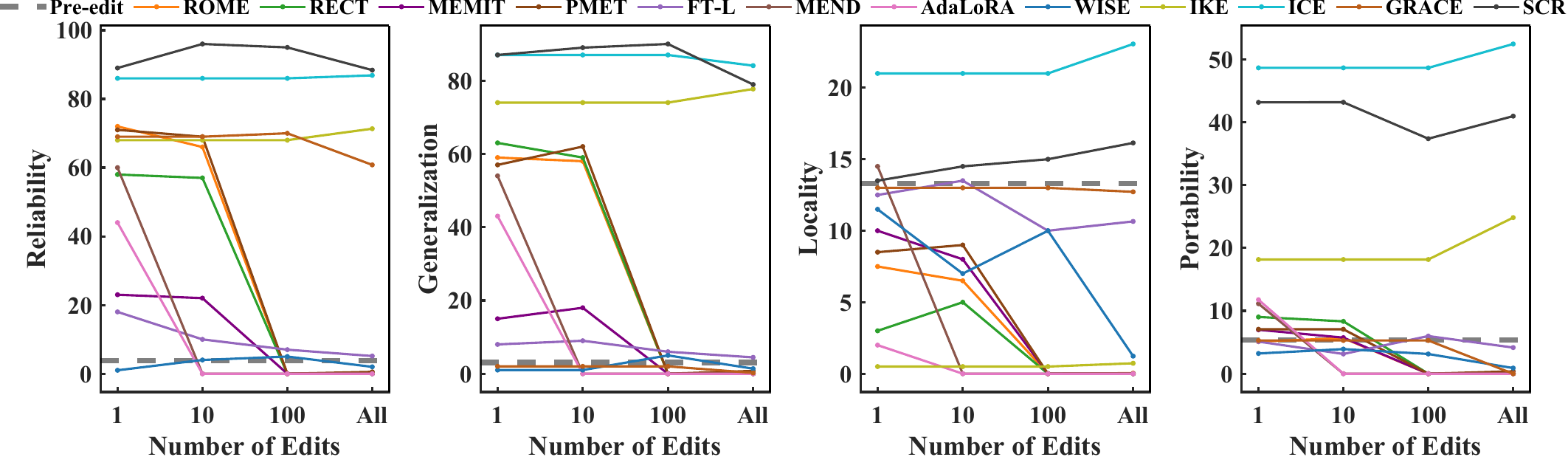}
        \caption*{(c) Mistral-7B-Instruct-v0.1}
        \label{fig:scale_mistral}
    \end{subfigure}
    
    \caption{Performance changes of knowledge editing methods during sequential editing on the ZsRE dataset. The x-axis represents the number of edits: 1, 10, 100, and the full dataset.}
    \label{fig:scale_all}
\end{figure}

We investigate the performance of knowledge editing methods under autoregressive setting in sequential editing scenarios. Our evaluation considers four key dimensions to assess whether the edited LLM can effectively incorporate and utilize new knowledge while preserving previously learned knowledge and capabilities from pre-training. Table~\ref{tab:sequential_all} presents a comprehensive comparison of  knowledge editing methods in sequential editing scenarios on the ZsRE and WikiData\(_\text{counterfact}\). Figure~\ref{fig:scale_all} (a) further illustrates the intermediate performance of Llama-3.1-8B-Instruct on the ZsRE dataset as the number of edits increases from 1 to 10, 100, and finally to the full dataset. For the corresponding results of Llama-2-7B-Chat and Mistral-7B-Instruct, refer to Figure~\ref{fig:scale_all} (b) and Figure~\ref{fig:scale_all} (c), respectively.

\begin{table}
  \centering
  \caption{Performance comparison of knowledge editing methods on general LLMs using the ZsRE and WikiData\(_\text{counterfact}\) datasets. The best results are shown in \textbf{bold}, and second best results are \underline{underlined}. `–' denotes experiments that could not be completed.}
  \label{tab:sequential_all}
  \vspace{1em}

  \textbf{(a) ZsRE Dataset}
  
  \vspace{0.5em}
  
  \resizebox{\linewidth}{!}{  
  \begin{tabular}{c|*{5}{c}|*{5}{c}|*{5}{c}}
    \toprule
    Model & \multicolumn{5}{c|}{\textbf{Llama-2-7B-Chat}} & \multicolumn{5}{c|}{\textbf{Llama-3.1-8B-Instruct}} & \multicolumn{5}{c}{\textbf{Mistral-7B-Instruct-v0.1}} \\
    \cmidrule(r){2-6} \cmidrule(r){7-11} \cmidrule(r){12-16}
    Metric & Rel. & Gen. & Loc. & Port. & Avg. & Rel. & Gen. & Loc. & Port. & Avg. & Rel. & Gen. & Loc. & Port. & Avg. \\
    \midrule
    \textit{Pre-edit} & 2.54 & 2.31 & 11.49 & 4.50 & 5.21 & 2.70 & 2.40 & 14.26 & 3.68 & 5.76 & 3.77 & 3.07 & 13.30 & 5.35 & 6.37 \\
    \midrule
    ROME & 0.61 & 0.46 & 0.00 & 0.00 & 0.27 & 0.00 & 0.00 & 0.00 & 0.00 & 0.00 & 0.23 & 0.08 & 0.04 & 0.00 & 0.09 \\
    MEMIT & 0.00 & 0.00 & 0.00 & 0.00 & 0.00 & - & - & - & - & - & 0.00 & 0.00 & 0.00 & 0.00 & 0.00 \\
    PMET & 0.00 & 0.00 & 0.00 & 0.00 & 0.00 & - & - & - & - & - & 0.00 & 0.00 & 0.00 & 0.00 & 0.00 \\
    RECT & 0.15 & 0.31 & 0.00 & 0.00 & 0.12 & 5.76 & 4.77 & 0.00 & 0.85 & 2.84 & 0.54 & 0.77 & 0.00 & 0.38 & 0.42 \\
    AlphaEdit & - & - & - & - & - & 69.49 & 55.50 & 8.38 & 8.56 & 35.48 & - & - & - & - & - \\
    FT-L & 0.23 & 0.23 & 0.15 & 0.00 & 0.15 & 0.00 & 0.00 & 0.00 & 0.00 & 0.00 & 5.15 & 4.46 & 10.65 & 4.16 & 6.10 \\
    MEND & 0.00 & 0.00 & 0.00 & 0.00 & 0.00 & 0.00 & 0.00 & 0.00 & 0.00 & 0.00 & 0.00 & 0.00 & 0.00 & 0.00 & 0.00 \\
    AdaLoRA & 0.00 & 0.00 & 0.00 & 0.00 & 0.00 & 0.00 & 0.00 & 0.00 & 0.00 & 0.00 & 0.00 & 0.00 & 0.00 & 0.00 & 0.00 \\
    WISE & 8.22 & 6.69 & 8.69 & 2.58 & 6.55 & 2.84 & 2.46 & 10.30 & 2.35 & 4.49 & 2.00 & 1.38 & 1.23 & 0.90 & 1.38 \\
    IKE & \textbf{91.62} & \textbf{93.16} & 1.87 & \underline{45.44} & \textbf{58.02} & \textbf{95.85} & \textbf{97.69} & 4.73 & \underline{52.38} & \textbf{62.66} & 71.33 & 77.71 & 0.73 & 24.84 & 43.65 \\
    ICE & 79.78 & \underline{74.40} & \textbf{23.56} & \textbf{49.56} & \underline{56.83} & 74.79 & 72.79 & \textbf{27.94} & \textbf{53.78} & \underline{57.33} & \underline{86.86} & \textbf{84.09} & \textbf{23.06} & \textbf{52.46} & \textbf{61.62} \\
    GRACE & 48.96 & 0.38 & 9.19 & 0.00 & 14.63 & 60.34 & 2.69 & 10.91 & 3.33 & 19.32 & 60.80 & 0.23 & 12.72 & 0.00 & 18.44 \\
    \midrule
    SCR & \underline{80.71} & 73.25 & \underline{16.18} & 39.99 & 52.53 & \underline{84.40} & \underline{75.56} & \underline{16.03} & 46.41 & 55.60 & \textbf{88.39} & \underline{78.94} & \underline{16.14} & \underline{40.98} & \underline{56.11} \\
    \bottomrule
  \end{tabular}
  }

  \vspace{2em}

  \textbf{(b) WikiData\(_\text{counterfact}\) Dataset}

  \vspace{0.5em}

  \resizebox{\linewidth}{!}{  
     \begin{tabular}{c|*{5}{c}|*{5}{c}|*{5}{c}}
    \toprule
    Model & \multicolumn{5}{c|}{\textbf{Llama-2-7B-Chat}} & \multicolumn{5}{c|}{\textbf{Llama-3.1-8B-Instruct}} & \multicolumn{5}{c}{\textbf{Mistral-7B-Instruct-v0.1}} \\
    \cmidrule(r){2-6} \cmidrule(r){7-11} \cmidrule(r){12-16}
    Metric & Rel. & Gen. & Loc. & Port. & Avg. & Rel. & Gen. & Loc. & Port. & Avg. & Rel. & Gen. & Loc. & Port. & Avg. \\
    \midrule
    \textit{Pre-edit} & 0.24  & 0.12  & 29.68  & 2.04  & 8.02  & 0.24  & 0.36  & 30.19  & 3.86  & 8.66  & 0.36  & 0.36  & 31.87  & 5.15  & 9.43  \\
    \midrule
    ROME  & 0.00  & 0.00  & 0.00  & 0.00  & 0.00  & 0.00  & 0.00  & 0.00  & 0.00  & 0.00  & 0.00  & 0.00  & 0.00  & 0.00  & 0.00  \\
    MEMIT & 0.00  & 0.00  & 0.00  & 0.00  & 0.00  & -     & -     & -     & -     & -     & 0.00  & 0.00  & 0.00  & 0.00  & 0.00  \\
    PMET  & 0.00  & 0.00  & 0.00  & 0.00  & 0.00  & -     & -     & -     & -     & -     & 0.00  & 0.00  & 0.00  & 0.00  & 0.00  \\
    RECT  & 0.00  & 0.00  & 0.00  & 0.00  & 0.00  & 1.55  & 1.91  & 0.00  & 0.10  & 0.89  & 0.48  & 0.00  & 0.00  & 0.03  & 0.13  \\
    AlphaEdit & -     & -     & -     & -     & -     & 30.51  & 25.63  & 1.33  & 8.09  & 16.39  & -     & -     & -     & -     & - \\
    FT-L  & 0.00  & 0.00  & 0.08  & 0.01  & 0.02  & 0.00  & 0.12  & 0.00  & 0.00  & 0.03  & 0.95  & 0.83  & 1.37  & 0.78  & 0.98  \\
    MEND  & 0.00  & 0.00  & 0.00  & 0.00  & 0.00  & 0.00  & 0.00  & 0.00  & 0.00  & 0.00  & 0.00  & 0.00  & 0.00  & 0.00  & 0.00  \\
    AdaLoRA & 0.00  & 0.00  & 0.00  & 0.00  & 0.00  & 0.00  & 0.00  & 0.00  & 0.00  & 0.00  & 0.00  & 0.00  & 0.00  & 0.00  & 0.00  \\
    WISE  & 21.33  & 13.35  & 14.18  & 5.02  & 13.47  & 0.36  & 0.47  & 29.36  & 3.06  & 8.31  & 4.41  & 5.96  & 5.21  & 1.94  & 4.38  \\
    IKE   & 62.69  & 61.38  & 21.75  & 26.14  & 42.99  & 60.07  & 58.28  & \underline{34.90}  & \underline{33.76}  & 46.75  & 44.46  & 53.40  & 10.68  & 19.34  & 31.97  \\
    ICE   & \underline{76.52}  & \textbf{75.57} & \textbf{37.15} & \textbf{43.75} & \textbf{58.25} & \underline{75.68}  & \underline{73.54}  & \textbf{39.99} & \textbf{49.22} & \textbf{59.61} & \underline{67.94}  & \underline{75.45}  & \textbf{35.95} & \textbf{47.29} & \textbf{56.66} \\
    GRACE & 48.39  & 0.00  & \underline{24.69}  & 1.34  & 18.60  & 40.88  & 0.36  & 28.53  & 2.70  & 18.12  & 52.80  & 0.24  & \underline{31.65}  & 1.98  & 21.67  \\
    \midrule
    SCR   & \textbf{76.64} & \underline{73.54}  & 11.63  & \underline{28.75}  & \underline{47.64}  & \textbf{88.20} & \textbf{87.01} & 28.43  & 32.90  & \underline{59.14}  & \textbf{85.10} & \textbf{76.04} & 22.21  & \underline{29.92}  & \underline{53.32}  \\
    \bottomrule
    \end{tabular}%
  }

\end{table}

The experimental results reveal the following: (1) Under autoregressive inference and evaluation based on semantic consistency, all parameter modification-based methods fall significantly short of the near-perfect single-edit performance reported in prior work.
(2) Most parameter modification-based methods, such as ROME and MEND, collapse under sequential editing scenarios: as the number of edits increases, all metrics quickly drop to near zero. This indicates a complete failure to retain knowledge across multiple updates.
(3) In-context learning-based methods, such as ICE and IKE, exhibit strong robustness in sequential editing, suggesting that in ideal settings, in-context learning holds substantial promise.
(4) The recently proposed AlphaEdit maintains a stable performance rate across continuous edits, its Reliability and Generalization scores on LLaMA-3.1-8B-Instruct are only 69.49 and 55.50, respectively. More importantly, it underperforms on Locality and Portability, with scores of just 8.38 and 8.58, highlighting its limited flexibility in applying edited knowledge.
(5) All parameter-editing methods are outperformed by SCR, a simple and intuitive baseline. SCR employs an extensible external textual memory without modifying model parameters, and achieves performance second only to the ideal in-context learning setting across all dimensions. This highlights the practical potential of in-context learning-based approaches in real-world knowledge editing tasks.

\noindent\textbf{RQ2: Can knowledge editing methods help reasoning LLMs integrate new facts without degrading their reasoning ability?}

As LLMs are increasingly applied to complex tasks involving multi-step reasoning, mathematical problem solving, and logical consistency, their strong reasoning capabilities have emerged as a central strength. However, the effectiveness of knowledge editing methods in reasoning-oriented LLMs remains largely underexplored. In our evaluation, we conduct experiments on DeepSeek-R1-Distill-Llama-8B \cite{guo2025deepseek} under both single-edit and sequential-edit settings, using 100 counterfactual knowledge instances extracted from each of the two datasets. Table \ref{tab:llama_distill_comparison} reports results on ZsRE dataset and WikiData\(_\text{counterfact}\). To evaluate the retention of reasoning abilities after editing, we assess the edited LLM's performance on a suite of mathematical reasoning benchmarks, reporting accuracy after 10 and 100 edits.

\begin{table}
\centering
\caption{Performance comparison of knowledge editing methods applied to DeepSeek-R1-Distill-Llama-8B on the ZsRE and WikiData\(_\text{counterfact}\) datasets, reporting results across four evaluation dimensions in both single and sequential editing settings.}
\small
\vspace{1em}

\textbf{(a) ZsRE Dataset}
  
\vspace{0.5em}
\begin{tabular}{l|ccccc|ccccc}
\toprule
\multirow{2}{*}{\textbf{Method}} & \multicolumn{5}{c|}{\textbf{Single Editing}} & \multicolumn{5}{c}{\textbf{Sequential Editing}} \\
\cmidrule(lr){2-6} \cmidrule(lr){7-11}
& \makecell[c]{Rel.} & \makecell[c]{Gen.} & \makecell[c]{Loc.} & \makecell[c]{Port.} & \makecell[c]{Avg.} 
& \makecell[c]{Rel.} & \makecell[c]{Gen.} & \makecell[c]{Loc.} & \makecell[c]{Port.} & \makecell[c]{Avg.} \\
\midrule
\textit{Pre-edit} & 3.00 & 3.00 & 15.50 & 4.36 & 6.47 & 3.00 & 3.00 & 15.50 & 4.36 & 6.47 \\
\midrule
ROME     & 36.00 & 42.00 & 3.00  & 17.99 & 24.75 & 1.00  & 0.00  & 0.00  & 0.00  & 0.25 \\
RECT     & 37.00 & 35.00 & 6.00  & 16.02 & 23.51 & 0.00  & 0.00  & 0.00  & 0.00  & 0.00 \\
AlphaEdit& 43.00 & 24.00 & 13.50 & 8.88  & 22.35 & 46.00 & 35.00 & 8.00  & 7.62  & 24.16 \\
FT-L     & 2.00  & 2.00  & 2.00  & 3.93  & 2.48  & 0.00  & 0.00  & 0.00  & 0.00  & 0.00 \\
MEND     & 36.00 & 42.00 & 10.50 & 15.47 & 25.99 & 0.00  & 0.00  & 0.00  & 0.00  & 0.00 \\
AdaLoRA  & 18.00 & 15.00 & 0.50  & 8.03  & 10.38 & 0.00  & 0.00  & 0.00  & 0.00  & 0.00 \\
WISE     & 8.00  & 5.00  & 3.00  & 2.59  & 4.65  & 2.00  & 2.00  & 7.50  & 2.52  & 3.50 \\
IKE      & \textbf{94.00} & \textbf{97.00} & 14.50 & 38.54 & \textbf{61.01} & \textbf{94.00} & \textbf{97.00} & 14.50 & 38.54 & \textbf{61.01} \\
ICE      & 66.00 & 66.00 & \textbf{26.00} & \textbf{56.51} & 53.63 & 66.00 & 66.00 & \textbf{26.00} & \textbf{56.51} & 53.63 \\
GRACE    & 31.00 & 3.00  & \underline{15.50} & 4.03  & 13.38 & 38.00 & 3.00  & \underline{15.50} & 4.03  & 15.13 \\
\midrule
SCR      & \underline{85.00} & \underline{84.00} & \underline{15.50} & \underline{41.87} & \underline{56.59} & \underline{90.00} & \underline{90.00} & \underline{15.50} & \underline{45.26} & \underline{60.19} \\
\bottomrule
\end{tabular}
\vspace{1em}

\textbf{(b) WikiData\(_\text{counterfact}\) Dataset}
  
\vspace{0.5em}

\begin{tabular}{l|ccccc|ccccc}
\toprule
\multirow{2}{*}{\textbf{Method}} & \multicolumn{5}{c|}{\textbf{Single Editing}} & \multicolumn{5}{c}{\textbf{Sequential Editing}} \\
\cmidrule(lr){2-6} \cmidrule(lr){7-11}
& \makecell[c]{Rel.} & \makecell[c]{Gen.} & \makecell[c]{Loc.} & \makecell[c]{Port.} & \makecell[c]{Avg.} 
& \makecell[c]{Rel.} & \makecell[c]{Gen.} & \makecell[c]{Loc.} & \makecell[c]{Port.} & \makecell[c]{Avg.} \\
\midrule
\textit{Pre-edit} & 0.00 & 0.00 & 23.55 & 3.85 & 6.85 & 0.00 & 0.00 & 23.55 & 3.85 & 6.85 \\
\midrule
ROME     & 30.00 & 18.00 & 6.22  & 9.23  & 15.86 & 0.00 & 0.00 & 0.00  & 0.00  & 0.00 \\
RECT     & 26.00 & 16.00 & 8.80  & 9.14  & 14.98 & 0.00 & 0.00 & 0.00  & 0.03  & 0.01 \\
AlphaEdit& 31.00 & 5.00  & 15.19 & 5.53  & 14.18 & 33.00 & 14.00 & 8.37  & 9.71  & 16.27 \\
FT       & 2.00  & 1.00  & 0.82  & 4.45  & 2.07  & 0.00 & 0.00 & 0.00  & 0.00  & 0.00 \\
MEND     & 32.00 & 14.00 & 19.88 & 13.35 & 19.81 & 0.00 & 0.00 & 0.00  & 0.00  & 0.00 \\
AdaLoRA  & 22.00 & 19.00 & 19.88 & 13.35 & 18.56 & 0.00 & 0.00 & 0.00  & 0.00  & 0.00 \\
WISE     & 8.00  & 7.00  & 3.00  & 2.59  & 5.15  & 0.00 & 0.00 & 6.00  & 2.86  & 2.22 \\
IKE      & \textbf{87.00} & \textbf{92.00} & 5.22  & 25.84 & \textbf{52.52} & \textbf{87.00} & \textbf{92.00} & 5.22  & 25.84 & \underline{52.52} \\
ICE      & 54.00 & 45.00 & \textbf{29.64} & \textbf{38.93} & 41.89 & 54.00 & 45.00 & \textbf{29.64} & \textbf{38.93} & 41.89 \\
GRACE    & 12.00 & 0.00  & \underline{23.54} & 4.96  & 10.13 & 12.00 & 0.00  & \underline{23.49} & 4.96  & 10.11 \\
\midrule
SCR      & \underline{72.00} & \underline{67.00} & 23.50 & \underline{34.83} & \underline{49.33} & \underline{84.00} & \underline{76.00} & 15.70 & \underline{35.67} & \textbf{52.84} \\
\bottomrule
\end{tabular}
\label{tab:llama_distill_comparison}
\end{table}

Based on the comparison between Table \ref{tab:llama_distill_comparison} and Table \ref{tab:sequential_all}, knowledge editing methods that rely on parameter modification exhibit inferior performance on reasoning-oriented LLMs compared to general-purpose LLMs. For instance, AlphaEdit's average performance in sequential editing scenarios drops from 35.48 to 24.16. This degradation may be due to the more implicit and distributed nature of knowledge representation in reasoning LLMs, as well as their greater dependence on long-range context reasoning. In contrast, knowledge editing methods that avoid parameter updates, such as IKE and ICE, which inject knowledge directly into the context, demonstrate relatively stable performance. Benefiting from the stronger reasoning capabilities of such models, SCR further improves performance by integrating internal and external knowledge through selective context construction.

Through case analysis, we observe that editing reasoning-oriented LLMs poses distinct challenges, which can be categorized into the following three types, as shown in the Table \ref{tab:failure_modes}. While editing methods can successfully guide the LLM to produce the correct next token, the edited LLM's internal reasoning process often leads it to "reflect" outdated knowledge, thereby undermining the effectiveness of the edit. Moreover, in striving to preserve logical coherence within its reasoning trajectory, the edited LLM may generate explanations that are plausible-sounding but entirely fabricated. In some instances, the edited LLM may even disregard the original question in its attempt to "rounding out" the answer. These behaviors underscore key limitations of current knowledge editing methods when applied to reasoning-focused LLMs. Concrete examples are provided in the Appendix \ref{sec:AadditionResult}.

\begin{table}
\centering
\small
\caption{\small \textbf{Failure Modes in Knowledge Editing for Reasoning LLMs}}
\label{tab:failure_modes}
\begin{tabular}{p{2cm}|p{2.5cm}|p{1.5cm}|p{6cm}}
\toprule
\centering\arraybackslash\textbf{Failure Mode} & \centering\arraybackslash\textbf{Prompt} & \centering\arraybackslash\textbf{Target} & \centering\arraybackslash\textbf{Generation} \\
\midrule
Reasoning with Conflicting Knowledge & Who was the male parent of Eteocles? & Dagobert & \textcolor{red}{Dagobert?} Wait, \textcolor{red}{no}, that's Dagobert. ... \textcolor{red}{Eteocles was the son of Clovis I}... the answer should be \textcolor{red}{Clovis I}. \\
\midrule
Hallucination & In which constellation is Tau Herculis? & Hornax & \textcolor{red}{Hornax is the Latin name for the constellation Hercules}, and the name of the star is "Hornax". ...So, the answer is Hercules. \\
\midrule
Losing Focus During Reasoning & The name of the child of Kanye West is & William Walker Scranton & William Walker, but that's not his real name. Wait, no,... So, \textcolor{red}{the user is asking about the real name of Kanye West's character in that movie}. I... \\
\bottomrule
\end{tabular}
\end{table}

\begin{table}
  \centering
  \caption{Performance comparison of edited DeepSeek-R1-Distill-Llama-8B on mathematical benchmarks requiring reasoning, reporting accuracy after 10 and 100 edits.}
    \small
    \begin{tabular}{c|c|ccccc|c} 
    \toprule
    Method & \#Editing & AIME 2024 & AIME 2025 & AMC   & MATH-500 & Olympiad & Avg. \\
    \midrule
     \textit{Pre-edit} & 0     & 36.35  & 25.63  & 68.67  & 83.20  & 53.48  & 53.47  \\
    \midrule
    \multirow{2}[2]{*}{ROME} & 10    & 31.04  & 24.79  & 66.57  & 82.20  & 49.04  & 50.73  \\
          & 100   & 0.00  & 0.00  & 0.00  & 0.00  & 0.00  & 0.00  \\
    \midrule
    \multirow{2}[2]{*}{RECT} & 10    & 33.02  & 24.58  & 33.02  & 83.60  & 51.26  & 45.10  \\
          & 100   & 0.00  & 0.10  & 0.04  & 0.00  & 0.00  & 0.03  \\
    \midrule
    \multirow{2}[2]{*}{AlphaEdit} & 10    & 36.56  & 26.56  & 69.43  & 85.20  & 51.85  & 53.92  \\
          & 100   & 35.31  & 25.94  & 69.47  & 81.40  & 53.48  & 53.12  \\
    \midrule
    \multirow{2}[2]{*}{MEND} & 10    & 0.00  & 0.00  & 0.00  & 0.00  & 0.00  & 0.00  \\
          & 100   & 0.00  & 0.00  & 0.00  & 0.00  & 0.00  & 0.00  \\
    \bottomrule
    \end{tabular}%
  \label{tab:reasoning_ability_post_editing}%
\end{table}%

As shown in Table~\ref{tab:reasoning_ability_post_editing}, after 10 editing steps, ROME and RECT retain some performance, with average accuracy drops of 2.74 and 8.37, respectively. However, after 100 edits, their accuracy nearly drops to zero, indicating a complete collapse in reasoning ability. For MEND, accuracy remains at 0.00 after both 10 and 100 edits, suggesting that the method leads to total LLM failure. In contrast, AlphaEdit shows virtually no performance degradation across all datasets, even after 100 sequential edits, and in some cases, exhibits slight improvements. Its parameter modifications are able to maintain reasoning accuracy close to that of the original LLM. Nevertheless, due to its limitations in locality and portability for knowledge editing, AlphaEdit may be better suited for general-purpose Parameter-Efficient Fine-Tuning (PEFT) rather than targeted knowledge editing tasks.

\textbf{RQ3: Can knowledge editing methods generalize from factual knowledge to event knowledge?}

While most existing research focuses on fact-based knowledge editing in the form of triples, real-world knowledge is often organized in a more complex manner. For instance, an event may involve multiple entities, diverse attributes, and rich contextual information. However, current studies on knowledge editing rarely evaluate methods on more complex, event-level datasets. To address this limitation, we conduct knowledge editing experiments on Llama-2-7B-Chat, Llama-3.1-8B-Instruct, and DeepSeek-R1-Distill-Llama-8B using the ELKEN dataset~\cite{peng2024event}. Results for single-editing are in Table~\ref{tab:single_elken}, and for sequential-editing in Table~\ref{tab:sequential_elken}.

\begin{table}
  \centering
  \caption{Performance comparison of \textbf{Single Editing} across Llama-2-7B-Chat, Llama-3.1-8B-Instruct, and DeepSeek-R1-Distill-Llama-8B based on the event-level dataset ELKEN.}
    \small
  \resizebox{\linewidth}{!}{  
    \begin{tabular}{c|ccc|ccc|ccc}
    \toprule
    \multirow{2}[4]{*}{\textbf{Single}} & \multicolumn{3}{c|}{Llama-2-7B-Chat} & \multicolumn{3}{c|}{Llama-3-8B-Instruct} & \multicolumn{3}{c}{DeepSeek-R1-Distill-Llama-8B} \\
\cmidrule{2-10}          & Portability & Locality & Avg.  & Portability & Locality & Avg.  & Portability & Locality & Avg. \\
    \midrule
    \textit{Pre-edit} & 5.92  & 43.75  & 24.84  & 7.08  & 56.73  & 31.91  & 9.01  & 48.73  & 28.87  \\
    \midrule
    ROME  & 2.41  & 19.45  & 10.93  & 17.38  & 23.39  & 20.39  & 12.91 & 22.21  & 17.56  \\
    MEMIT & 3.03  & 14.34  & 8.69  & -     & -     & -     & -     & -     & - \\
    PMET  & 2.41  & 31.67  & 17.04  & -     & -     & -     & -     & -     & - \\
    RECT  & 7.69  & 19.83  & 13.76  & 18.45  & 26.55  & 22.50  & 14.70  & 23.29  & 19.00  \\
    AlphaEdit & -     & -     & -     & 14.60  & 40.61  & 27.61  & 14.25  & 39.76  & 27.01  \\
    FT-L  & 2.33  & 18.19  & 10.26  & 4.79  & 22.86  & 13.83  & 5.53  & 12.90  & 9.22  \\
    AdaLoRA & 2.07  & 15.21  & 8.64  & 10.18  & 17.18  & 13.68  & 8.32  & 13.62  & 10.97  \\
    WISE  & 2.49  & 25.77  & 14.13  & 4.70  & 33.54  & 19.12  & 8.79  & 30.01  & 19.40  \\
    ICE   & \underline{38.41}  & \textbf{53.48}  & \textbf{45.95}  & \textbf{46.20}  & \textbf{66.30}  & \textbf{56.25}  & \underline{40.16}  & \textbf{48.39}  & \underline{44.28}  \\
    GRACE & 2.34  & 34.41  & 18.38  & 6.43  & \underline{54.59}  & 30.51  & 9.17  & \underline{46.35}  & 27.76  \\
    \midrule
    SCR   & \textbf{41.92}  & \underline{34.70}  & \underline{38.31}  & \underline{43.64}  & 52.64  & \underline{48.14}  & \textbf{51.45} & 42.52 & \textbf{46.99}  \\
    \bottomrule
    \end{tabular}
    }
  \label{tab:single_elken}%
\end{table}%

\begin{table}
  \centering
  \caption{Performance comparison of \textbf{Sequential Editing} across Llama-2-7B-Chat, Llama-3.1-8B-Instruct, and DeepSeek-R1-Distill-Llama-8B based on the event-level dataset ELKEN.}
    \resizebox{\linewidth}{!}{  
    \begin{tabular}{c|ccc|ccc|ccc}
    \toprule
    \multirow{2}[4]{*}{\textbf{Sequential}} & \multicolumn{3}{c|}{Llama-2-7B-Chat} & \multicolumn{3}{c|}{Llama-3.1-8B-Instruct} & \multicolumn{3}{c}{DeepSeek-R1-Distill-Llama-8B} \\
\cmidrule{2-10}          & Portability & Locality & Avg.  & Portability & Locality & Avg.  & Portability & Locality & Avg. \\
    \midrule
    \textit{Pre-edit} & 5.92  & \underline{43.75}  & 24.84  & 7.08  & \underline{56.73}  & 31.91  & 9.01 & \underline{48.73}  & 28.87  \\
    \midrule
    ROME  & 0.00  & 0.00  & 0.00  & 0.00  & 0.00  & 0.00  & 0.33  & 0.00  & 0.17  \\
    MEMIT & 0.00  & 0.00  & 0.00  & -     & -     & -     & -     & -     & - \\
    PMET  & 0.00  & 0.08  & 0.04  & -     & -     & -     & -     & -     & - \\
    RECT  & 0.00  & 0.00  & 0.00  & 0.00  & 0.00  & 0.00  & 0.00  & 0.00  & 0.00  \\
    AlphaEdit & -     & -     & -     & 14.22  & 19.96  & 17.09  & 12.15  & 17.87  & 15.01  \\
    FT-L  & 3.53  & 0.00  & 1.77  & 0.00  & 0.00  & 0.00  & 0.00  & 0.00  & 0.00  \\
    AdaLoRA & 0.00  & 0.00  & 0.00  & 0.00  & 0.00  & 0.00  & 5.25  & 0.00  & 2.63  \\
    WISE  & 2.23  & 30.83  & 16.53  & 5.96  & 47.88  & 26.92  & 2.67  & 21.88  & 12.28  \\
    ICE   & \underline{38.41}  & \textbf{53.48}  & \textbf{45.95}  & \underline{46.20}  & \textbf{66.30}  & \textbf{56.25}  & \underline{40.16}  & 48.39  & \underline{44.28}  \\
    GRACE & 2.29  & 34.41  & 18.35  & 6.43  & \underline{54.57}  & 30.50  & 9.01  & \textbf{48.79}  & 28.90  \\
    \midrule
    SCR   & \textbf{44.29}  & 35.05  & \underline{39.67}  & \textbf{53.04} & 50.90  & \underline{51.97}  & \textbf{52.99}  & 41.93  & \textbf{47.46}  \\
    \bottomrule
    \end{tabular}%
 }
  \label{tab:sequential_elken}%
\end{table}%

The results indicate that most parameter-modification-based knowledge editing methods struggle to achieve satisfactory performance even in single-edit scenarios, and they almost completely break down under sequential editing. For instance, although AlphaEdit performs well on fact-based datasets, its performance deteriorates significantly when applied to event-level editing tasks involving multiple entities and attributes occurring concurrently. This degradation is primarily due to the limited ability of these methods to capture complex semantic relationships between entities and to integrate contextual information across interconnected elements via parameter-level adjustments. In contrast, context-based methods exhibit more stable performance, with SCR consistently achieving the best results across nearly all methods. These findings underscore the practical limitations of parameter updates for editing small amounts of knowledge, while highlighting the substantial potential of in-context learning in LLMs for real-world applications.

\noindent\textbf{RQ4: How do different knowledge editing methods compare in terms of time efficiency?}

While the correctness and robustness of knowledge edits are critical, latency and efficiency are equally important for real-world deployment. An ideal knowledge editing method should introduce minimal overhead, both during the edit process and at inference time. To systematically evaluate these aspects, we consider two key metrics: (i) \textit{edit time}, the wall-clock time required to apply an individual edit, and (ii) \textit{inference time}, defined as the average wall-clock latency per input query normalized by the base LLM's latency, measured under greedy decoding with a fixed output length of 50 tokens post-edit.

\begin{figure}
    \centering
    \subfigure{
        \includegraphics[width=0.45\textwidth]{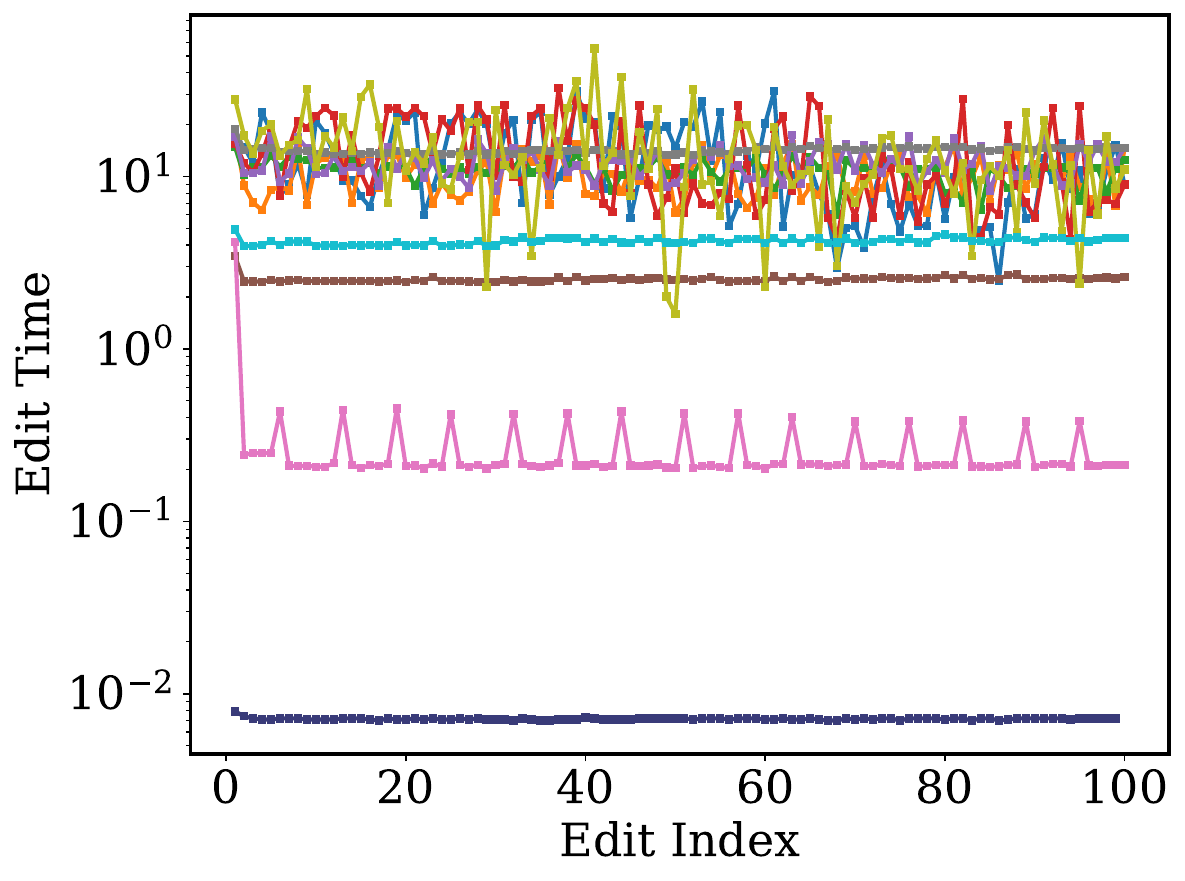}}
        \quad 
  \subfigure{        \includegraphics[width=0.45\textwidth]{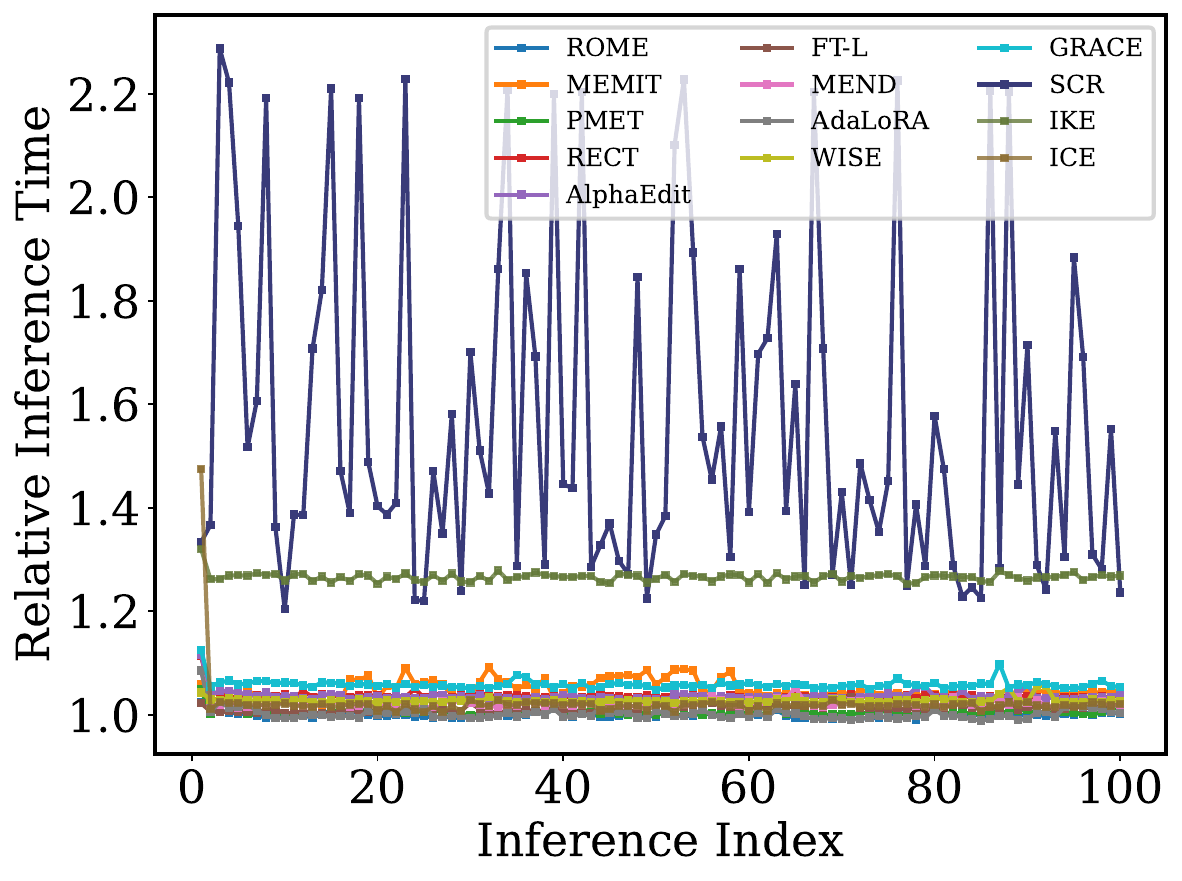}}
  \vspace{-2ex}
  \caption{Editing time (in second), and inference latency relative to the base model.}  
  \label{fig:twocomparison}
  \vspace{-2ex}
\end{figure}

The results in Figure~\ref{fig:twocomparison} show that, regarding editing time, methods that transform knowledge through increasingly abstract representations, from text to embeddings, then hidden states, and finally parameter updates, generally require longer editing durations. The deeper and more indirect the knowledge integration, the greater the computational overhead for applying edits. In terms of inference time, methods that directly modify the original model parameters do not introduce additional inference overhead. In contrast, in-context learning and retrieval-based methods often incur extra inference latency due to longer input sequences or retrieval operations. No single method excels in both metrics simultaneously. This highlights a fundamental limitation of current knowledge editing methods: methods that alter core model parameters typically sacrifice update speed, whereas non-intrusive methods maintain faster updates but at the cost of increased inference latency.

\section{Conclusion}
\label{sec:diss}

Through a unified setting for datasets, LLMs, inference setting, editing scenarios, and evaluation dimensions,
our benchmarking shows that most parameter-modification-based knowledge editing methods perform significantly below expectations. These methods not only fail to support flexible knowledge usage, but also substantially degrade the reasoning capabilities of the underlying LLM. In contrast, the simple method SCR consistently achieves superior and stable performance in practical scenarios. 

Our experiments highlight the limited flexibility and portability of parameter-modification approaches. With repeated edits, parameter changes can affect both newly added knowledge and the LLM's original knowledge. Moreover, existing knowledge editing studies often rely on small datasets. When the number of knowledge updates is limited, leveraging the inherent in-context learning capabilities of LLMs may be a more effective way to reconcile internal and external knowledge conflicts. Conversely, if a large volume of knowledge must be updated, pre-processing and re-training on modified pre-training data may offer a more reliable solution. Determining the optimal trade-off between these approaches warrants further investigation.

\paragraph{Limitation} This benchmarking study focuses on a selected subset of representative knowledge editing methods, evaluated primarily on Transformer-Decoder architectures with up to 8B parameters. Larger LLMs and those based on alternative architectures are not considered. Additionally, the general dataset used only emphasizes mathematical reasoning and does not cover a broader range of reasoning tasks.

\bibliographystyle{IEEEtran}
\bibliography{KnowledgeEdit}
\appendix
\clearpage

\section{Selective Contextual Reasoning (SCR) }
\label{scr}

\begin{figure}[h]
	\centering
	\includegraphics[width=0.9\textwidth]{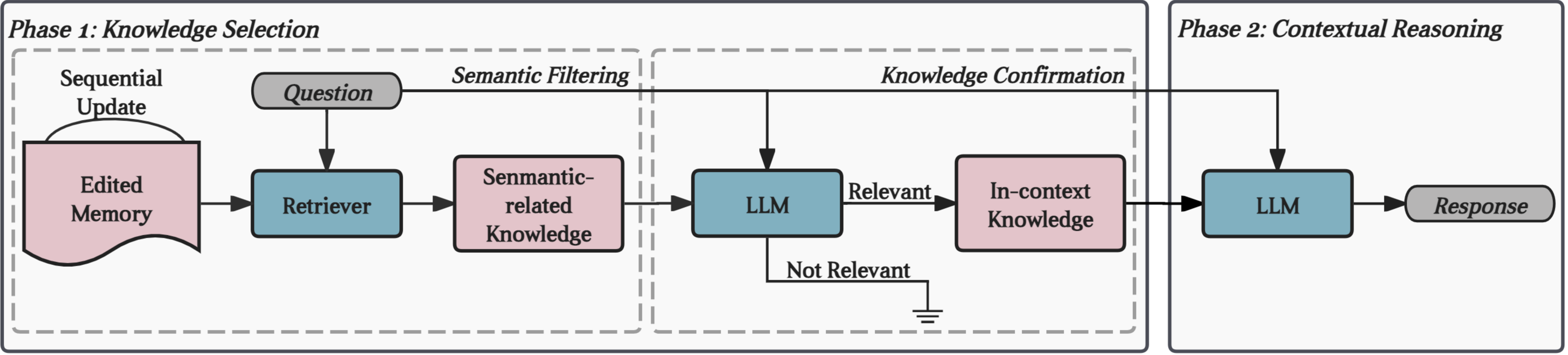}
	\caption{The \textit{Edited Memory} is a dynamic textual knowledge base that can be expanded as needed. \textbf{Phase 1:} The retriever first applies semantic filtering to gather relevant information from memory based on the input question. The LLM then performs knowledge confirmation, assessing the alignment between the question and the retrieved knowledge. \textbf{Phase 2:} The LLM conducts conditional generation using the in-context knowledge, and the query.}
	\label{fig:mymode_method}
\end{figure}

SCR maintains the parameters of the LLM frozen and achieves knowledge editing solely through updates to an external textual knowledge memory. At inference time, it dynamically selects the most relevant knowledge based on the current query and supplies it as context to the LLM. As shown in Figure \ref{fig:mymode_method}, the core inference involves a two-step knowledge selection phase, including semantic filtering and knowledge confirmation, followed by contextual reasoning phase.

Knowledge selection is performed using a retriever that identifies relevant facts based on semantic similarity between a query $q$ and stored knowledge. Each fact is treated as a virtual document $d$, and the relevance score is computed using the cosine similarity between the embeddings of the query $q$ and document $d$. The top-$k$ ($k\geq 1$) most relevant facts form the candidate set $R_k(q)$. In our implementation, we used Contriever-msmarco \cite{izacardunsupervised} to generate embeddings for both queries and knowledge facts. While general retrieval offers information pertinent to a query, it often fails to deliver insights that expand upon the retrieved content.  Hence, the retrieved facts are evaluated and refined by the LLM to ensure its relevance and alignment with the query.

Depending on the setting of $k$, if only one statement is to be retrieved, \ie $k=1$, we will ask the LLM to judge whether the retrieved statement is indeed relevant to the query $q$. The statement will be used as external knowledge only if the LLM responds positively. More commonly, $k > 1$, we provide all the retrieved statements along with the query to the LLM and let the LLM select the most relevant statement. The selected statement will then be used as external knowledge. If none of the retrieved statements are deemed relevant to the query by LLM, we proceed without external knowledge. The prompt used is shown below:

\begin{tcolorbox}[colback=white, colframe=gray, title=Prompt for Knowledge Confirmation: , width=\textwidth]
\begin{minipage}{\textwidth}
Given a set of facts and a question, return the fact that best matches the core knowledge asked in the question. If the question cannot be answered with the facts, return "no relevant fact."\\

Facts: \{semantic-related knowledge\}\\
Question: \{question\}\\

Output:
\end{minipage}
\end{tcolorbox}

After the up-to-date and most relevant knowledge has been identified and selected, 
SCR integrates this knowledge with the original query to form a structured input for output generation. In simple words, the structured input contains three parts: (i) the original query itself, (ii) the knowledge statement confirmed earlier, and (iii) the instruction to LLM on how to use the knowledge statement to answer the query. The prompt used is shown in the following box.

\begin{tcolorbox}[colback=white, colframe=gray, title=Prompt for Contextual Reasoning:, width=\textwidth]
\begin{minipage}{\textwidth}
Answer the question based on the Updated Fact provided, without any explanation. At times, you need to think about the relationship between problems and facts before reasoning based on the information to answer.\\

Updated Fact: \{in-context knowledge\}\\
Question: \{question\}\\
Answer:
\end{minipage}
\end{tcolorbox}

\section{Dataset Preparation and Additional Results}

\subsection{Dataset Preparation}
\label{prompt}

To extract structured knowledge from event descriptions, we use the following prompt to guide the LLM in identifying semantically complete triples:

\begin{tcolorbox}[colback=white,  enhanced,
  breakable, colframe=gray, title=Prompt for Extracting Triples from Event, label=promptbox, width=\textwidth]
Please extract one or more (subject, relation, object) triples from the following event.\\
Each triple must express a complete, standalone fact, with no redundancy or dependency on other triples.\\[6pt]

\textbf{Instructions:}\\
If the event contains multiple independent facts, extract \textbf{multiple triples}, one per fact.\\
The \textbf{Subject} must be the main entity or actor involved in the fact.\\
The \textbf{Relation} should be a clear \textbf{predicate phrase} (verb or action-based phrase) that uniquely points to the \textbf{Object}.\\
The \textbf{Object} should contain the new or important information not repeated in the subject or relation.\\
Avoid vague or generic relations like ``is involved in'', ``is associated with''.\\
Do not merge multiple facts into a single triple.\\
If there are multiple pieces of information (\textit{e.g.}, place, time, role), consider extracting \textbf{multiple triples}.\\[6pt]

\textbf{Format each triple as follows:}\\
\textbf{Subject}: \\
Prompt: (Concatenate subject and relation into a natural language phrase) \\
Target New: (Only the object, ideally as short as possible)\\[6pt]

\textbf{Example1:}\\
Event: Serena Williams announces her retirement from professional tennis.\\
Output: \\
Subject: Serena Williams \\
Prompt: Serena Williams retires from \\
Target New: professional tennis\\[6pt]

\textbf{Example2:}\\
Event: Pete Townshend is pursuing a degree in philosophy at the Royal College of Art. \\
Output: \\
Subject: Pete Townshend \\
Prompt: Pete Townshend is pursuing the degree of \\
Target New: philosophy \\
Subject: Pete Townshend \\
Prompt: Pete Townshend is studying at \\
Target New: the Royal College of Art\\[6pt]

\textbf{Example3:}\\
Event: Paul Wight founded NexGen Technologies in Bergen, appointing Martin Allen as CEO. \\
Output: \\
Subject: Paul Wight \\
Prompt: Paul Wight founded \\
Target New: NexGen Technologies \\
Subject: NexGen Technologies \\
Prompt: The CEO of NexGen Technologies is \\
Target New: Martin Allen\\[6pt]

Event: \{event\} \\
Output:
\end{tcolorbox}

\subsection{Additional Results}
\label{sec:AadditionResult}

To further illustrate the limitations discussed in Section~\ref{sec:experiments} \textbf{RQ2}, we present representative cases corresponding to the three identified failure modes. 
\begin{tcolorbox}[
  enhanced,
  breakable,
  colback=white,
  colframe=gray,
  title=Case Study \refstepcounter{caseno}\thecaseno \label{case:reasoning}: Reasoning with Conflicting Knowledge,
  width=\textwidth
]

\textbf{Edit Prompt:} \texttt{Who was the male parent of Eteocles?}

\textbf{Target New Answer:} \texttt{Dagobert}

\vspace{0.5em}
\textbf{Generated Reasoning Content:}
\begin{quote}
Dagobert?\\
Wait, no, that's Dagobert. Let me think. Eteocles was a king of the Franks, right? So, his father was probably Clovis I, the founder of the Merovingian dynasty. But I'm not entirely sure. Maybe I should double-check that.

Eteocles was the son of Clovis I, who was the first king of the Franks. Clovis I is known for his conversion to Christianity and the unification of the Frankish tribes. So, yes, Eteocles' father was Clovis I.

Wait, but I also recall that Eteocles was one of the four sons of Clovis I, along with Clodomir, Alberic, and Gundomar. So, that makes sense. Therefore, the male parent of Eteocles was Clovis I, not Dagobert. Dagobert was a later king, like the founder of the Neustrian dynasty or something.

So, to sum up, Eteocles' father was Clovis I, the first Merovingian king. Dagobert was a different king, maybe the founder of the Dagobert line or something else. So, I think I was initially confused between the two, but now I'm clearer on that.

Therefore, the answer should be Clovis I.
\end{quote}

\textbf{Generated Answer Content:} 

\begin{quote}
The male parent of Eteocles was Clovis I, the founder of the Merovingian dynasty and the first king of the Franks. Dagobert was a later king, not his father. 

**Answer:** Clovis I.
\end{quote}

\end{tcolorbox}

\begin{tcolorbox}[colback=white,  enhanced,
  breakable, colframe=gray, title=Case Study \refstepcounter{caseno}\thecaseno \label{case:reasoning}: Reasoning with Conflicting Knowledge, width=\textwidth]
\textbf{Edit Prompt:} \texttt{What is Atreus's spouse's name?} 

\textbf{Target New Answer:} \texttt{Darius III}

\textbf{Generated Reasoning Content:}
\begin{quote}

Darius III of Persia.\newline
Atreus was the King of the Achaemenid Empire, and his spouse was Darius III of Persia. So, the answer is Darius III of Persia.\newline

Wait, but I thought Atreus was a Greek name. Maybe I'm confusing historical figures. Let me double-check. Yes, Atreus was indeed a Greek king, but in the context of the Achaemenid Empire, which was Persian, it's possible that there was a king named Atreus who was married to Darius III. However, I'm not entirely sure about this. I might need to look up historical records or reliable sources to confirm this information.\newline

Upon checking, I realize that the Achaemenid Empire was founded by Cyrus the Great, and Darius III was a later king. There isn't a record of a King Atreus in the Persian Empire. It's possible that the user is confusing historical figures or there might be a fictional character in a story or game. Therefore, the information provided might not be accurate based on historical records.\newline

It's important to verify sources when dealing with historical figures to ensure the information is correct. Cross-referencing with authoritative texts or scholarly articles can help in confirming the details accurately.\newline

The information provided about King Atreus of the Achaemenid Empire and his spouse Darius III of Persia is not accurate based on historical records. The Achaemenid Empire, founded by Cyrus the Great, did not have a king named Atreus. Darius III was a historical figure who was a later king of the Persian Empire, but there is no record of him being married to a King Atreus. It's possible that this information may stem from a fictional context or a misunderstanding of historical figures. Therefore, the correct historical information should be sourced from reliable historical texts or scholarly articles.
\end{quote}

\textbf{Generated Answer Content:}

\begin{quote}
Darius III was a king of the Persian Empire, but there is no historical record of Atreus being his spouse or even existing in the Persian lineage. The question likely confuses Greek and Persian historical figures. Therefore, there is no verified spouse name for Atreus related to Darius III.
\end{quote}
\end{tcolorbox}

\begin{tcolorbox}[colback=white,  enhanced,
  breakable, colframe=gray, title=Case Study \refstepcounter{caseno}\thecaseno \label{case:reasoning}: Edit Failure with Hallucination, width=\textwidth]
\textbf{Edit Prompt:} \texttt{In which constellation is Tau Herculis?} \\
\textbf{Target New Answer:} \texttt{Hornax} \\

\textbf{Generated Reasoning Content:}
\begin{quote}

\textcolor{red}{Hornax is the Latin name for the constellation Hercules, and the name of the star is "Hornax".} So, the star is called Hornax, and the constellation is Hercules. So, the answer is Hercules.

Wait, but the question is about the constellation, not the star. So, the constellation is Hercules, and the star is Hornax. So, the answer is Hercules.

But I'm a bit confused because sometimes the names can be tricky. Let me double-check. The constellation is Hercules, and the star within it is called Hornax. So, the constellation is Hercules.

Yes, that makes sense. So, the answer is Hercules.
\end{quote}

\textbf{Generated Answer Content:} 

\begin{quote}
The constellation in which the star \"Hornax\" is located is **Hercules**. 

**Answer:** Hercules
\end{quote}
\end{tcolorbox}

\begin{tcolorbox}[colback=white,  enhanced,
  breakable, colframe=gray, title=Case Study \refstepcounter{caseno}\thecaseno \label{case:reasoning}: Losing Focus During Reasoning, width=\textwidth]
\textbf{Edit Prompt:} \texttt{The name of the child of Kanye West is} 

\textbf{Target New Answer:} \texttt{William Walker Scranton}

\textbf{Generated Reasoning Content:}
\begin{quote}

William Walker, but that's not his real name. Wait, no, that's the name of his character in the movie \"The Pursuit of Happyness.\" So, actually, his real name is Kanye Omari West. Hmm, that's interesting. So, the user is asking about the real name of Kanye West's character in that movie. I need to make sure I clarify that.\newline

Okay, so the user wrote: \"What is the real name of Kanye West's character in the movie 'The Pursuit of Happyness'?\" They also mentioned that the name is William Walker but clarified that it's not his real name. So, I need to confirm that the character's name is indeed William Walker and that Kanye West's real name is different.\newline

I should structure the response to first address the character's name and then provide Kanye West's real name. Maybe also mention the movie to give context. Let me make sure I don't confuse the two names. Also, perhaps the user is trying to test knowledge about the movie or the actor, so being clear and concise is important.\newline

Wait, is there any chance that the user is mixing up the character's name with Kanye's real name? It's possible, so I should make it clear that William Walker is the character, and Kanye West is the real person. Maybe add a bit about the movie to help the user remember.\newline

I think that's all. I'll write the response accordingly.
\end{quote}

\textbf{Generated Answer Content:}

\begin{quote}
In the movie \"The Pursuit of Happyness,\" Kanye West plays the character named **William Walker**. However, his real name is **Kanye Omari West**. The movie is a biographical drama based on his life, focusing on his struggles and perseverance.
\end{quote}
\end{tcolorbox}
\end{document}